\pdfoutput=1
\documentclass[preprint,12pt,authoryear]{elsarticle}

\usepackage{amssymb}
\usepackage{amsmath}
\usepackage{times}
\usepackage{soul}
\usepackage{url}
\usepackage[hidelinks]{hyperref}
\usepackage[utf8]{inputenc}
\usepackage[small]{caption}
\usepackage{graphicx}
\usepackage{amsmath}
\usepackage{amsthm}
\usepackage{booktabs}
\usepackage{algorithm}
\usepackage{algorithmic}
\usepackage[switch]{lineno}
\usepackage[defaultcolor=green]{changes}

\usepackage{rotating}
\usepackage{amssymb}
\usepackage{diagbox}
\usepackage{enumitem}
\usepackage{subfigure}
\usepackage{subfiles}
\usepackage{multirow}
\usepackage{booktabs} 
\usepackage{xcolor}
\usepackage{fancyhdr}
\usepackage{hyperref}

\newtheorem{lemma}{Lemma}
\newtheorem{definition}{Definition}

\newtheorem{myrule}{Rule}

\journal{arXiv}

\begin{document}

\begin{frontmatter}

\title{An Identifiable Cost-Aware Causal Decision-Making Framework Using Counterfactual Reasoning} 

\author[1]{Ruichu Cai\corref{corresponding}}
\cortext[corresponding]{Corresponding author.}
\ead{cairuichu@gmail.com}

\author[1]{Xi Chen}
\author[1]{Jie Qiao}
\author[1]{Zijian Li}
\author[1]{Yuequn Liu}
\author[1]{Wei Chen}
\author[2]{Keli Zhang}
\author[2]{Jiale Zheng}

\affiliation[1]{organization={School of Computer Science, Guangdong University of Technology}, 
city={Guangzhou}, 
postcode={510006}, 
country={China}}
\affiliation[2]{organization={Huawei Noah's Ark Lab,Huawei}, 
city={Shenzhen}, 
postcode={518116}, 
country={China}}

\begin{abstract}
Decision making under abnormal conditions is a critical process that involves evaluating the current state and determining the optimal action to restore the system to a normal state at an acceptable cost. However, in such scenarios, existing decision-making frameworks highly rely on reinforcement learning or root cause analysis, resulting in them frequently neglecting the cost of the actions or failing to incorporate causal mechanisms adequately. By relaxing the existing causal decision framework to solve the necessary cause, we propose a minimum-cost causal decision (MiCCD) framework via counterfactual reasoning to address the above challenges.
Emphasis is placed on making counterfactual reasoning processes identifiable in the presence of a large amount of mixed anomaly data, as well as finding the optimal intervention state in a continuous decision space. Specifically, it formulates a surrogate model based on causal graphs, using abnormal pattern clustering labels as supervisory signals. This enables the approximation of the structural causal model among the variables and lays a foundation for identifiable counterfactual reasoning. With the causal structure approximated, we then established an optimization model based on counterfactual estimation. The Sequential Least Squares Programming (SLSQP) algorithm is further employed to optimize intervention strategies while taking costs into account. Experimental evaluations on both synthetic and real-world datasets reveal that MiCCD outperforms conventional methods across multiple metrics, including F1-score, cost efficiency, and ranking quality(nDCG@k values), thus validating its efficacy and broad applicability.
\end{abstract}

\end{frontmatter}

\section{Introduction}

Decision making is a fundamental process for problems-solving and goal achievement, which involves evaluating the current state and selecting the most appropriate action from a set of alternatives while maintaining acceptable cost \cite{yuan2024partial,shehadeh2024expert}. Although decision-making has demonstrated effectiveness across various domains \cite{waqar2024intelligent,hu2019deep,fang2022framework,schmitt2023automated,venkatesan2024data,saadatmand2024evaluation,zheng2024survey}, its role becomes even more critical when dealing with anomalies, where timely and accurate decisions can fix system failures, reduce losses, and improve operational efficiency. In high-stakes sectors like industry and healthcare, poor decisions—especially under anomalies—may lead to severe, costly, or even unethical outcomes. These challenges highlight the urgent need for robust decision-making frameworks tailored to handle anomalies.

\begin{figure}[h]  
    \centering  
    \includegraphics[width=1\textwidth]{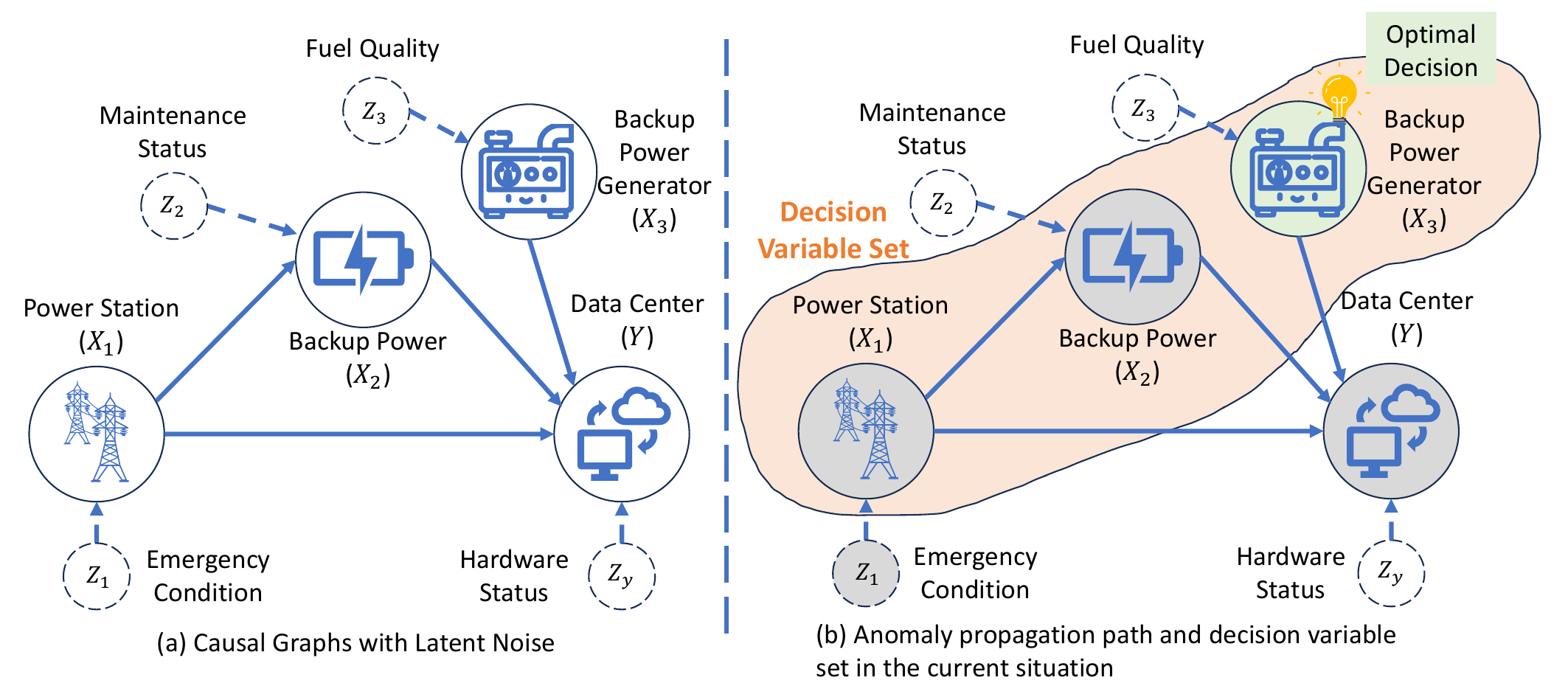} 
    \caption{ A toy decision-making example. If a data center ($Y$) loses power due to an earthquake ($Z_1$) and relies on backup power ($X_2$), RCA may recommend repairing the power station (the root cause ($X_1$)), but this is not feasible because the backup power ($X_2$) will be exhausted before the repair is complete. A feasible, lower-cost solution—such as increasing the backup generator’s fuel supply ($X_3$)—requires reasoning beyond root cause identification, incorporating cost and counterfactual evaluation. Here, solid nodes are the observed variables and dashed nodes are the potential noise. Grey fills in nodes represent anomalies, and green represents the optimal decision when anomalies occur.}
    \label{fig:intro}  
\end{figure}

Current approaches to decision making under anomalies largely fall into two categories: reinforcement learning (RL) and root cause analysis (RCA).  
RL methods predominantly emphasize optimal learning actions through a trial-and-error process from online \cite{mnih2015human,schulman2015trust,schulman2017proximal} or offline \cite{wu2019behavior,kumar2020conservative,kostrikov2022offline} learning. However, such exploratory strategies often entail substantial costs and may raise ethical concerns. 
RCA methods, on the other hand, aim to identify the underlying causes of abnormal behavior, enabling decision-making based on the inferred root cause. Correlation-based RCA methods \cite{ribeiro2018anchors,ide2021anomaly,casalicchio2019visualizing,carletti2019explainable,pan2020unsupervised,hyun2025collaboration,tamanampudi2020data,du2017deeplog}, risk identifying spurious dependencies due to a lack of causal grounding, while causal RCA \cite{rehak2023counterfactual,assaad2023root,budhathoki2022causal,nguyen2024root} often recommend actions solely based on root causes, without accounting for feasibility or cost in the current state. A critical challenge remains: not just diagnosing the anomaly, but identifying effective, feasible, and cost-efficient interventions. A typical example is illustrated in Figure \ref{fig:intro}(b).

To address the aforementioned gaps, we propose Minimum Cost Causal Decision (MiCCD), a novel decision-making framework that integrates causal modeling with cost-aware counterfactual reasoning to guide decision making under anomalies. Specifically, MiCCD formulates decisions as interventions and adaptively identifies cost-effective counterfactual intervention strategies, thereby eliminating the need for continuous trial-and-error processes. Given the necessity for counterfactual inference and the potentially large and continuous action space present in real-world applications, MiCCD is designed to address the following two critical challenges:
1) \textbf{Counterfactual identification:} How can we identify the counterfactual consequences given an abnormal state and a specific action?
2) \textbf{Cost-aware optimization:} How can we find the optimum minimum-cost decision within a large and continuous action space?

For the first challenge, MiCCD relies on accurately modeling causal mechanisms, including the distribution of exogenous noise terms (e.g., the effect of $X_1$ and $Z_2$ on $X_2$). Identifiability of these noise terms is essential; otherwise, counterfactual predictions become ambiguous, reducing decision reliability. MiCCD addresses this by disentangling noise patterns and clustering abnormal states, reconstructing their distinct distributional impacts through hierarchical causal graph decomposition.

For the second challenge, MiCCD balances necessity (whether an intervention resolves the anomaly) and economy (its cost). Using surrogate models to approximate causal effects, it iteratively estimates the impact of interventions and employs Sequential Least Squares Programming (SLSQP) to optimize intervention selection and intensity under a cost-minimization objective.

The key contributions of this work are summarized as follows:

\begin{enumerate}
    \item A novel cost-aware decision-making framework, MiCCD, is proposed, integrating counterfactual reasoning with intervention optimization to address anomaly resolution.
    \item A surrogate model is developed to approximate underlying causal relationships, enabling counterfactual estimation of intervention outcomes. The necessity and feasibility of interventions are quantified through a defined necessity probability and cost-aware formulation.
    \item MiCCD is validated through extensive experiments on both synthetic and real-world datasets, demonstrating superior performance and cost efficiency in comparison to existing methods.
\end{enumerate}

\section{Related Work}
\paragraph{RL Methods}

Reinforcement learning (RL) encompasses a diverse range of methods, each designed to learn from either online or offline trial-and-error processes, to solve decision-making tasks. In traditional RL, the decision policy is optimized through direct, online interactions with the environment (known as online learning), where the agent learns by actively exploring and exploiting the environment in real time. Typical methods include Deep Q-Networks (DQN) \cite{mnih2015human}, Trust Region Policy Optimization (TRPO) \cite{schulman2015trust}, Proximal Policy Optimization (PPO) \cite{schulman2017proximal}, and Soft Actor-Critic (SAC) \cite{haarnoja2018soft}. In contrast, offline RL focuses on learning policies from static trial-and-error datasets. Representative offline RL methods include Behavior Regularized Actor Critic (BRAC) \cite{wu2019behavior}, Conservative Q-Learning (CQL) \cite{kumar2020conservative}, Implicit Q-Learning (IQL) \cite{kostrikov2022offline}, and Batch-Constrained Q-learning (BCQ). However, RL-based models still rely on explicit or implicit data derived from trial-and-error processes to achieve effective decision-making.

\paragraph{RCA Methods}
Root cause analysis (RCA) methods can be categorized into correlation-based and causal approaches. Correlation-based methods include LIME \cite{ribeiro2018anchors}, LC \cite{ide2021anomaly}, and NaiveRCA \cite{strumbelj2010efficient}, which evaluate the importance of features; CAFCA \cite{hyun2025collaboration}, which analyzes data based on categorical groupings; and DeepLog \cite{du2017deeplog}, which establishes statistical associations between input features and target variables. But these methods often produce spurious dependencies due to their lack of causal reasoning. For example, high temperatures can cause machine damage and increase air humidity, but correlation-based methods might erroneously identify air humidity as the root cause.
Causal-based methods address these limitations by incorporating causal structures. 
For example, to locate the root cause, JRCS \cite{rehak2023counterfactual} identifies anomalies in a given causal graph, while CausalRCA \cite{budhathoki2022causal} and BIGEN \cite{nguyen2024root} use structured causal models and Bayesian reasoning to identify root causes. EEL \cite{strobl2023sample} uses structural equation models and Shapley-values for sample-specific RCA. However, these methods focus solely on identifying root causes without considering the cost or consequences of interventions, limiting their applicability in decision-making scenarios.

\paragraph{Causal-based Decision Methods}
Recent work by \cite{qin2024rehearsal} proposed a rehearsal learning framework to mitigate unexpected futures (AUF) using probabilistic graphical models and structural equations. \cite{duavoiding} extended this framework with AUF-MICNS, which incorporates decision costs and adapts to non-stationary environments. However, these methods operate at the population level rather than on individual-level decisions.

\section{Problem Formulation}
Decision with causality is based on specific causal structures, which are commonly represented by directed acyclic graphs (DAG) \cite{pearl2009causality} over variables. Consider random variables $\mathbf{X}=\left \{X_{1},\dots ,X_{d}\right \}$ with index set $V :=\left \{ 1,\dots ,d \right \} $. A graph $\mathcal{G}=(V,\varepsilon  ) $ consists of nodes $V$ and edges $\varepsilon\subseteq V ^{2}$ with $(i,j)  \in \varepsilon$ for any $i,j \in V$. A node $i$ is called a parent of $j$ if $e_{ij} \in \varepsilon$ and $e_{ji} \notin \varepsilon$. The set of parents of $j$ is denoted as $\mathbf{PA} _{j}^{\mathcal{G} } $. The graph representation of causality is formally discussed as follows:
\begin{definition}[Structural Causal Models]
\label{scm}
A structural causal model (SCM) $\mathfrak{C} :=(\mathbf{S },\mathbf{Z} )$ consists of a collection $\mathbf{S}$ of $d$ functions $X_{i}:=f_{i}(\mathbf{PA}_{i}^{\mathcal{G} } ,Z_{i})  $, $i\in \left [ d \right]$, where $\mathbf{PA}_{i}\subset \left \{ X_{1},\dots ,X_{d}\right \} \setminus \left \{ X_{i} \right \} $ are the parents of $X_{i}$, and a joint distribution $\mathbf{Z} =\left \{ Z_{1} ,\dots, Z_{d} \right \} $ over the noise variables, which are required to be jointly independent.
\end{definition}
Let $\left \{ {\mathbf{x}^{(i)} } \right \} _{i=1}^{n}$ and $\left \{ {y^{(i)} } \right \} _{i=1}^{n} $ denote the set of samples, where $\mathbf{x} =\left [ x_{1},\dots ,x_{d}   \right ] \in \mathbb{R}^{d}  $ and $y \in \mathbb{R}  $, with an additional target corresponding to the SCM.
Similarly, $\mathbf{z} =\left [ z_{1},\dots ,z_{d},z_{y}   \right ] \in \mathbb{R}^{d+1}  $ represents the sample of $\mathbf{Z}$. 

The objective of this study is to develop and formalize a decision-making framework aimed at addressing existing challenges in a causal system. In such systems, deviation expected behavior may arise either due to anomalies(e.g., sensor faults, adversarial attacks) or intentional modifications (e.g., policy adjustments, system upgrades), each introducing distinct noise patterns that significantly alter the distribution of counterfactual outcomes. 
Without loss of generality, the problem is modeled as an abnormality scenario, in which the abnormality is caused by a perturbation of the noise distribution, from $Z _{i}\sim \mathcal{N}(\mu_i,\sigma_i^2)$ to $Z' _{i}\sim \mathcal{N}(\mu_i',\sigma_i'^2)$. As illustrated in Figure \ref{fig:intro}(a) and (b), if an anomaly arises due to a shift in the distribution of $Z_1$, then $Z_1$ is part of the anomalous noise distribution. Conversely, if the anomaly is induced by noise in other variables, the distribution of $Z_1$ differs. This is subsequently propagated through the causal graph, resulting in altered system outcomes.

Decision-making is framed as a counterfactual problem, wherein the objective is to identify a variable or set of variables that, when subjected to intervention, can mitigate anomalies in the target variable $Y$ at minimal intervention cost. Anomaly resolution is achieved by classifying the target variable based on a threshold $t$. Specifically, if an anomaly is detected ($Y > t$), the target variable is assigned a value of 1; otherwise, a value of 0 denotes a resolved anomaly.

To rigorously assess the impact of an intervention on the target, the Probability of Necessity (PN) metric \cite{pearl2009causality} is introduced. In contrast to traditional approaches, PN quantifies the \textit{causal necessity} of an intervention within a counterfactual framework, and is formally defined as follows:
\begin{equation}
\begin{aligned}
    PN(X\to Y)=P(Y_{do(\mathbf{X} =\mathbf{ x^{*}} ) }=0|\mathbf{X}=\mathbf{x}  ,Y=1),
\end{aligned}
\end{equation}
Here, \( Y_{do(\mathbf{X} = \mathbf{x}^*)} \) represents the counterfactual outcome. Intuitively, PN seeks to answer the following question: \textit{What is the probability that the outcome $Y$ would be zero (i.e., resolved) if an intervention changes the original input $\mathbf{x}$ to $ \mathbf{x}^*$?}. This formulation provides a principled framework for decision makers to assess whether an intervention is genuinely necessary to effect change in the outcome.

The objective is thus reformulated as minimizing the intervention cost, subject to the PN exceeding a predefined threshold $\iota$:

\begin{equation}
    \begin{aligned}
        \mathbf{ x^{s}}  &= \arg\min_{\mathbf{x^{*}}  \in \mathbb{R}^{n} } C(do(\mathbf{X} =\mathbf{x^{*}}  ),\mathbf{x} ) \\
        &\text{s.t. } P(Y_{do(\mathbf{X} =\mathbf{x^{*}})  }=0|\mathbf{X} =\mathbf{x} ,Y=1) \ge \iota ,
    \end{aligned}
\end{equation}
Here,  $\mathbf{x}^s = \left [ x_1^s, \dots, x_d^s \right ] $ represents the optimal intervention vector that alters $Y$ from 1 to 0 at minimal cost. The cost function $C$ typically depends on factors such as the step size between $\mathbf{ x^{*}}$ and $\mathbf{ x}$, the intervention costs associated with individual variables, and the number of affected variables. The specific form of $C$ can vary depending on the application context.

\section{The Minimum Cost Causal Decision Framework}
In this section, the Minimum-Cost Causal Decision (MiCCD) framework is introduced, which is designed to address two critical challenges: (1) the identifiability of counterfactuals under causal structures, and (2) the optimization of cost-aware interventions to determine minimum-cost decisions.

\begin{figure*}[ht]
    \centering
    \includegraphics[width=\textwidth]{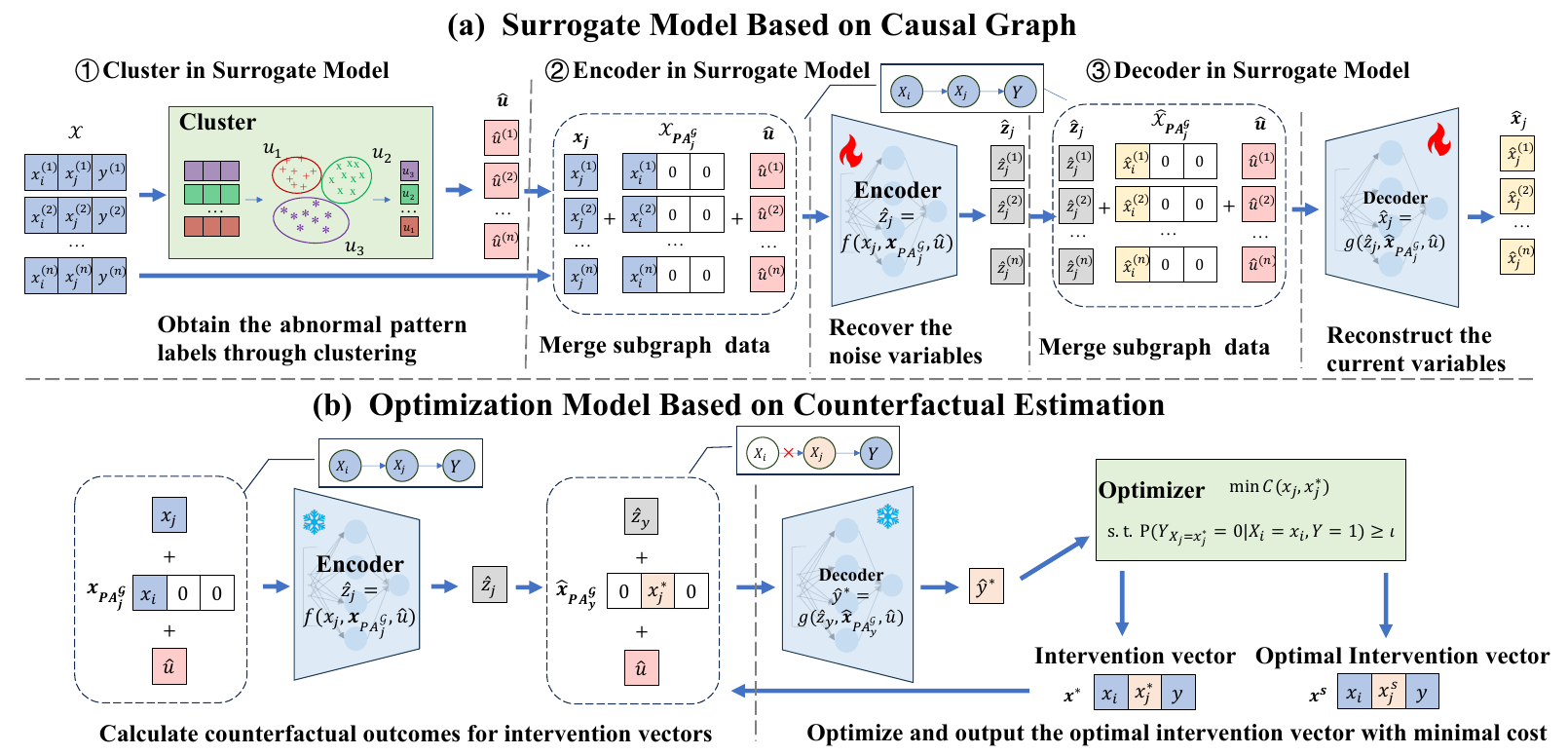}
    \caption{The MiCCD framework comprises two components: (a) a surrogate model that approximates the SCM by recovering noise variables and predicting outcomes to enable counterfactual reasoning, and (b) an optimization method to identify the minimum-cost intervention based on counterfactual reasoning. In the surrogate model, abnormal pattern labels are first obtained through clustering, and are subsequently incorporated into subgraph samples-- comprising the current variable and its parent variables—to recover the corresponding noise variables. These noise variables, along with the parent variables, are then fed into the decoder to reconstruct the current variable, or the counterfactual variable if a parent variable is subject to intervention. In the optimization phase, the encoder and decoder are employed to estimate counterfactual outcomes for candidate intervention vectors generated by the optimizer. Through iterative refinement, the framework determines the cost-minimizing intervention.
}
    \label{fig: model}
\end{figure*}

\subsection{Identifiability of Counterfactual Reasoning under Intervention}
One of the primary challenge in counterfactual reasoning is the difficulty of recovering the noise variables $ \mathbf{Z} $ without performing actual interventions. This challenge is addressed by first formalizing the intervention process, identifying the main difficulties associated with estimating post-intervention distributions, and followed by proposing a solution underpinned by theoretical guarantees.

In order to identify effective interventions, we begin by selecting the intervention vector $\mathbf{x}^*$. Not every interventions in $\mathbf{x}^*$ is effective in influencing the target variable $ Y $. The following rule is applied to filter out ineffective interventions:
\begin{myrule}
\label{ru}
For all elements in $\mathbf{X^{*}}=\mathbf{X}$, the following screening process is required:
\begin{itemize}
    \item S1. Remove all elements that satisfying $x_{i}^{*}=x_{i}$:
    \begin{equation}
        \mathbf{X^{*}_{R} } \gets \mathbf{X^{*} }\setminus \left \{ X_{i}^{*} | x_{i}^{*}= x_{i} ,\exists i\in \left \{ 1,\dots,i,\dots ,d \right \}   \right \} 
    \end{equation}
    \item S2. For each remaining element $X_{i}^{*} \in \mathbf{X^{*}_{R}} $, identify each path $E(X_{i}^{*},Y)$ from $X_{i}^{*}$ to $Y$. If for each path $e\in E( X_{i}^{*},Y)$, all intermediate elements in the path belong to the set of remaining elements $\mathbf{X^{*}_{R}}$, then delete $X_{i}^{*}$ from $\mathbf{X^{*}_{R}}$. Repeat the above steps in causal order until no further deletions can be made:
    \begin{equation}
        \mathbf{X^{*}_{R} } \gets \mathbf{X^{*}_{R} }\setminus \left \{ X_{i}^{*} | \forall e\in E( X_{i}^{*},Y),\forall m\in e,m\in \mathbf{X^{*}_{R}}  \right \} 
    \end{equation}
\end{itemize}
The actions $\mathbf{ x^{*}_{R}}$ corresponding to the final $\mathbf{X^{*}_{R}}$ are deemed effective.
\end{myrule}

This screening process ensures that only interventions capable of influencing the targets are considered, thereby reducing the search space. For example, in the case of  $X_2 \leftarrow X_1 \rightarrow Y$, $X_2$ is screened out as it does not influence $Y$.

To estimate the distribution of the target node $ Y $ after intervening on $\mathbf{X}^*_R$, the following expression can be derived based on Bayesian inference. Here, $\mathbf{DE}_{R}^{\mathcal{G} }$ represents the descendant set of the node set $R$ in $\mathcal{G}$:
\begin{equation}
\label{f1}
\begin{aligned}
&p(y \mid do(\mathbf{ X_{R}^{*}} = \mathbf{ x_{R}^{*}}))  = \int_{\mathbf{x}_{\mathbf{DE}_{R}^{\mathcal{G} }}}  \int_{\mathbf{z}_{\mathbf{DE}_{R}^{\mathcal{G} }}}  do(\mathbf{ X_{R}^{*}} = \mathbf{ x_{R}^{*}}) \\
 &\times \prod_{i \in \mathbf{DE}_{R}^{\mathcal{G} }}  p(x_{i}|\mathbf{x} _{\mathbf{PA}_{i}^{\mathcal{G} }},z _{i}) \times  \prod_{i \in \mathbf{DE}_{R}^{\mathcal{G} }}p(z_{i}) d\mathbf{x}_{\mathbf{DE}_{R}^{\mathcal{G} }}d\mathbf{z} _{\mathbf{DE}_{R}^{\mathcal{G} }}
\end{aligned}
\end{equation}

From Equation \ref{f1}, it is observed that estimating the post-intervention distribution requires the identification of the noise distribution $ \mathbf{Z} $ and the distribution of $\mathbf{DE}_{R}^{\mathcal{G} }$ post-intervention. This presents several challenges: (1) Real-world conditions rarely allow for trial-and-error approaches, necessitating a correct and directly derivable surrogate model; (2) In the absence of strong assumptions, the distribution of $ \mathbf{Z} $ in mixed abnormal data is inherently difficult to identify \cite{locatello2019challenging}; (3) Identifying the distribution of $ \mathbf{Z} $ may require the incorporation of auxiliary information from dataset clustering, though anomaly pattern labels are often unavailable.

To address the aforementioned challenges, different noise distributions are distinguished by clustering the generated anomaly labels. These labels are provided to the variational autoencoder as supervisory signals, guiding the encoder in recovering the noise variables (Lemma \ref{the1}). For instance, in the context of power supply, clustering can differentiate between the noise distributions of 'power station damage caused by an earthquake' and 'back-up power overload', ensuring the identifiability of noise recovery in the surrogate model.

\subsubsection{Identification of Clustering}
The identifiability of clustering is established under the following sufficient conditions:
\begin{lemma}[Sufficient Condition for Identifiability Based on Weakly Separable Variables. \cite{tahmasebi2018identifiability}]
\label{the1}
Any mixture model with latent variables $\theta =(F_{1:K};\mathbf{w} )\in \Theta _{K,L,M}$ such that $\mathcal{L}_{w} (F_{1:K};\mathbf{w} )\ge 2K$ is identifiable.
\end{lemma}
\paragraph{Implication} 
For $ (d+1) $-dimensional data generated according to Definition \ref{scm}, it is assumed that each sample contains at least one abnormal variable, and each variable, except the target variable, exhibits at least one abnormality. Under normal conditions, each variable $ X_i $ follows a Gaussian distribution $ \mathcal{N}(\mu_i, \sigma_i^2) $,  where $\mu _{i}$ and $\sigma _{i}^{2}$ represent the mean and variance of the variable $X_{i}$, respectively. Under abnormal conditions, $ X_i $ follows $ \mathcal{N}(\mu_i', \sigma_i'^2) $, affecting its descendants $ \mathbf{X} _{\mathbf{DE} _i^{\mathcal{G} }} $. For each variable, there exists at least one distinct distribution, either under normal and abnormal conditions or under its own abnormality and under the influence of ancestral abnormalities. Therefore, at least $ 2d=2K $ weakly separable variables satisfy the identifiable conditions.

This result guarantees that clustering methods can effectively distinguish abnormal patterns, thereby providing the necessary supervision for noise recovery.
\subsubsection{Identifiability of Latent Noise Variables.} 
The prior on the latent variables $p_{\mathbf{\theta } }(\mathbf{z}|\mathbf{u})$ is assumed to be conditionally factorial, where each element of $z_{i}\in \mathbf{z} $ follows a univariate exponential family distribution given the conditioning variable $\mathbf{u}$. The conditioning on $\mathbf{u}$ is performed via an arbitrary function $\lambda  (\mathbf{u} )$ (such as a search table or neural network), which outputs the individual parameters of the exponential family $\lambda_{i,j}$. The probability density function is therefore given by:
\begin{equation}
\label{zu}    p_{\mathbf{T},\mathbf{\lambda}} (\mathbf{z}|\mathbf{u}) = \prod_{i} \frac{Q_{i}(z_{i})}{Z_{i}(\mathbf{u})} \exp\left[\sum_{j=1}^{k} T_{i,j}(z_{i})\lambda _{i,j}(\mathbf{u})\right]
\end{equation}

The identifiability of latent noise variables $ \mathbf{Z} $ is established through the following lemma:
\begin{lemma}[Identifiability of Latent Noise Variables \cite{khemakhem2020variational}]

    Assume the observed data is generated according to Definition \ref{scm} and Equations \ref{zu}, with parameters $(\mathbf{f},\mathbf{T} ,\mathbf{\lambda}   )$. The identifiability of these parameters holds under the following conditions: \begin{itemize}
        \item A1: The mixing function $f_{i}$ in Definition \ref{scm} is injective.
        \item A2: The sufficient statistics $T_{i,j}$ in Equation \ref{zu} are differentiable almost everywhere, and $(T_{i,j})_{1\le j\le k}$ are linearly independent on any subset of $\mathbf{X}$ of measure greater than zero.
        \item A3: There exist $nk+1$ distinct points $\mathbf{u} ^{0} ,\cdots ,\mathbf{u} ^{nk} $ such that the matrix \begin{equation}
            L=(\lambda (\mathbf{u} _{1})-\lambda (\mathbf{u} _{0}),\cdots ,\lambda (\mathbf{u} _{nk})-\lambda (\mathbf{u} _{0}))
        \end{equation}
        of size $nk \times nk$ is invertible.
    \end{itemize}
    then the parameters $(\mathbf{f},\mathbf{T} ,\mathbf{\lambda}   )$ are $\sim _{A}$-identifiable.
\end{lemma}
\paragraph{Implication}Intuitively, this theoretical result demonstrates that latent variables $ \mathbf{Z} $ can be identified by leveraging their distributional variations, which correspond to different clusters as established in Lemma \ref{the1}. These clusters reflect distinct generative processes or abnormal patterns in the data. Consequently, this lemma offers the theoretical foundation for the recovery of $ \mathbf{Z} $, thereby ensuring the identifiability of counterfactuals in our framework.

\subsection{Optimization of the Minimum Cost Causal Decision}

In the previous section, we established the theoretical guarantees for counterfactual identifiability, highlighting the necessity of auxiliary labels to distinguish noise distributions and facilitate accurate counterfactual reasoning. Building upon this foundation, this section presents the optimization produce of the MiCCD framework. As illustrated in Figure \ref{fig: model}, MiCCD comprises two principal components: (a) a surrogate model for counterfactual reasoning, which leverages clustering and a causally identifiable variational autoencoders to approximate the SCM, and (b) an optimization scheme designed to identify the minimum-cost intervention vector that effectively mitigates anomalies in the target variable with high PN.

\subsubsection{Surrogate Model Based on Causal Graph}
The surrogate model, depicted in Figure \ref{fig: model}(a), is designed to perform abduction by recovering latent noise variables and predicting endogenous variables through a neural architecture based on variational inference. While the model inherently supports both abduction and prediction, it also enables intervention, thereby facilitating counterfactual reasoning. This is achieved by applying interventions to the observed data and leveraging the recovered noise variables to generate updated predictions (as discussed in Section \ref{sec:opt}). The surrogate model is composed of three key components: (1) an anomaly pattern clustering module that provides auxiliary labels to guide noise recovery, (2) an encoder that infers the structured noise variables, and (3) a decoder that performs prediction based on the learned causal structure.

\paragraph{Anomaly Pattern Cluster} To generate the auxiliary variable, a key characteristic of abnormal data is exploited: each anormaly pattern is characterized by distinct statistical properties, which give rise to separable clusters in the data distribution.
Specifically, the Gaussian Mixture Model (GMM) \cite{dempster1977maximum} is employed for clustering, under the assumption that the data are generated from a mixture of Gaussian components. Pattern labels $\mathbf{u}$ are assigned to the abnormal samples through the optimization of GMM parameters. These labels serve as control signals that guide noise recovery, offering critical information regarding the sources of anomalies and thus supporting the identifiability required for accurate counterfactual inference.

\paragraph{Variational Autoencoder with Embedded Causal Structure} A variational autoencoder (VAE) framework is adopted by the surrogate model to recover latent noise variables and reconstruct endogenous variables, thereby facilitating an accurate approximation of the SCM. To ensure alignment with the underlying causal structure, an SCM-informed Evidence Lower Bound (ELBO) is derived and utilized as the training objective. The formal expression of this objective is provided in Equation \ref{elbo}, and the detailed derivation is included in \ref{app:elbo}.
\begin{equation}
\label{elbo}
\begin{aligned}
    \log p(\mathbf{x} |\mathbf{u})
=& \sum_{j=1}^{d}\underbrace{\mathbb{E} _{q(z_j|\mathbf{x}_{PA_{j}^{\mathcal{G} }},x_j,\mathbf{u} )}\left [ \log p(x_j|z_j,\mathbf{x}_{PA_{j}^{\mathcal{G} }},\mathbf{u} ) \right ]}_{\mathcal{L}_{RE} }  \\ &-\sum_{j=1}^{d}D_{KL}( q(z_j|\mathbf{x}_{PA_{j}^{\mathcal{G} }},x_j,\mathbf{u} )\parallel p(z_j|\mathbf{x}_{PA_{j}^{\mathcal{G} }},\mathbf{u}))
\end{aligned}
\end{equation}

The term $\sum_{j=1}^{d}$ denotes a summation over individual nodes. The term  $\mathcal{L}_{RE}$ quantifies the reconstruction error, while $D_{KL}$ enforces consistency between the prior distribution $p(z_j|\mathbf{x}_{PA_{j}^{\mathcal{G} }},\mathbf{u})$ and the posterior distribution $q(z_j|\mathbf{x}_{PA_{j}^{\mathcal{G} }},x_j,\mathbf{u} )$.

Equation \ref{elbo} presents the optimization scheme for optimizing the encoder and decoder in the surrogate model, as depicted in Figure \ref{fig: model}(a). First, the observation data are processed through the anomaly clustering model to generate pattern labels. These labels are integrated with the input data to capture variations driven by abnormalities. Subsequently, since the current variable depends on its parent variables and its own noise, the encoder, which is implemented as an MLP, processes each variable, along with its parent variables and the corresponding abnormal pattern, to recover the latent noise variable $\mathbf{z}$, under the guidance of the causal order defined by the causal graph.

The decoder, which is also implemented as an MLP, utilizes the recovered noise variables and parent variables to reconstruct the target variable. The reconstructed variable then serves as the parent input for subsequent variables in a sequential manner. This iterative process ensures consistency with the underlying causal relationships, including both linear and nonlinear interactions. In particular, the encoder corresponds to the inverse structural equation responsible for noise recovery, whereas the decoder represents the forward structural equations, facilitating accurate predictions of endogenous variables. For a specific MLP training configuration, please refer to \ref{app:a5}.

\subsubsection{Optimization Model Based on Counterfactual Estimation}\label{sec:opt}
Building on the learned surrogate model, counterfactual reasoning is implemented by modifying its decoder and optimizing the intervention decision to achieve the minimum cost while satisfying the constraints of the target variable. This section outlines the counterfactual estimation process and the optimization framework used for identifying the minimum-cost intervention.

\paragraph{Counterfactual Estimation} Counterfactual reasoning is performed through three basic steps: abduction, intervention, and prediction. First, abduction involves recovering the latent noise variables \( z \) using the encoder component of the surrogate model to infer \( z \) from the observed data together with the anomaly pattern labels \( \mathbf{u} \). The intervention then applies a hypothetical scenario by performing a hard intervention \( do(X_i = x_i^*) \) on a variable \( X_i \), setting its value to \( x_i^* \), and removing its dependence on parent variables. Finally, prediction propagates the effects of intervention through the causal graph: for each child variable \( X_j \in X_{Chi_i} \), the encoder recovers its noise \( z_j \), and the decoder reconstructs its value using the intervened parent value \( x_i^* \) and the inferred noise \( z_j \). This process is repeated iteratively in causal order, resulting in the counterfactual outcome \( y^* \) for the target variable \( Y \). Together, these steps enable the surrogate model to simulate hypothetical scenarios and provide a foundation for optimizing intervention decisions.

\paragraph{Minimal Cost Optimization} The problem of identifying the optimal intervention vector \( \mathbf{x}^* \) that resolves the anomaly in the target variable \( Y \) with minimum intervention cost is now addressed. The goal is to identify the intervention vector that minimizes the cost function \( C(do(\mathbf{X} = \mathbf{x}^*), \mathbf{x}) \) subject to the constraint \( P(Y_{do(\mathbf{X} = \mathbf{x}^*)} = 0 \mid \mathbf{X} = \mathbf{x}, Y = 1) \ge \iota \). However, in many real-world scenarios, the action space for interventions is often large and continuous, with potentially infinitely many intervention vectors \( \mathbf{x}^* \) in this space. 

To address this problem, it is formulated as a constrained optimization task. The necessary conditions for optimality are derived using the Karush-Kuhn-Tucker (KKT) conditions. The Lagrangian function \( L \) is defined as follows:
\begin{equation}
\label{lagrange}
L(\mathbf{x}^*, \lambda) = C(do(\mathbf{X} = \mathbf{x}^*), \mathbf{x}) + \lambda \cdot g(\mathbf{x}^*),
\end{equation}
The constraint function is given by \( g(\mathbf{x}^*) = \iota - P(Y_{do(\mathbf{X} = \mathbf{x}^*)} = 0 \mid \mathbf{X} = \mathbf{x}, Y = 1) \), and the corresponding Lagrange multiplier is \( \lambda \geq 0 \).

To efficiently solve this non-linear optimization problem, the Sequential Least Squares Programming (SLSQP) algorithm \cite{boggs1995sequential} is employed. An iterative search is conducted by SLSQP to identify the optimal intervention vector \( \mathbf{x}^* \) and the associated Lagrange multiplier \( \lambda \), such that the KKT conditions are satisfied. Simultaneously, the constraints of cost minimization and $PN\ge \iota$ are enforced. Here, PN functions as a 'validity filter', restricting the search space to intervention solutions with a high probability of success, even in large continuous action spaces. Through the integrating counterfactual reasoning and numerical optimization, the proposed framework  facilitates efficient and cost-sensitive decision-making in complex real-world scenarios.

\section{Experiments}
In this section, the datasets, baseline methods, and evaluation metrics employed in the experiments are first described. Subsequently, numerical results and in-depth discussions are provided to address the following key research questions: Q1. How does MiCCD perform in generalization settings compared to existing baselines? Q2. How effective is root cause analysis when guided by intervention-based strategies? Q3. To what extent does each component of the MiCCD framework contribute to counterfactual reasoning?
\begin{figure}[htbp]
    \centering
    \includegraphics[width=0.7\textwidth]{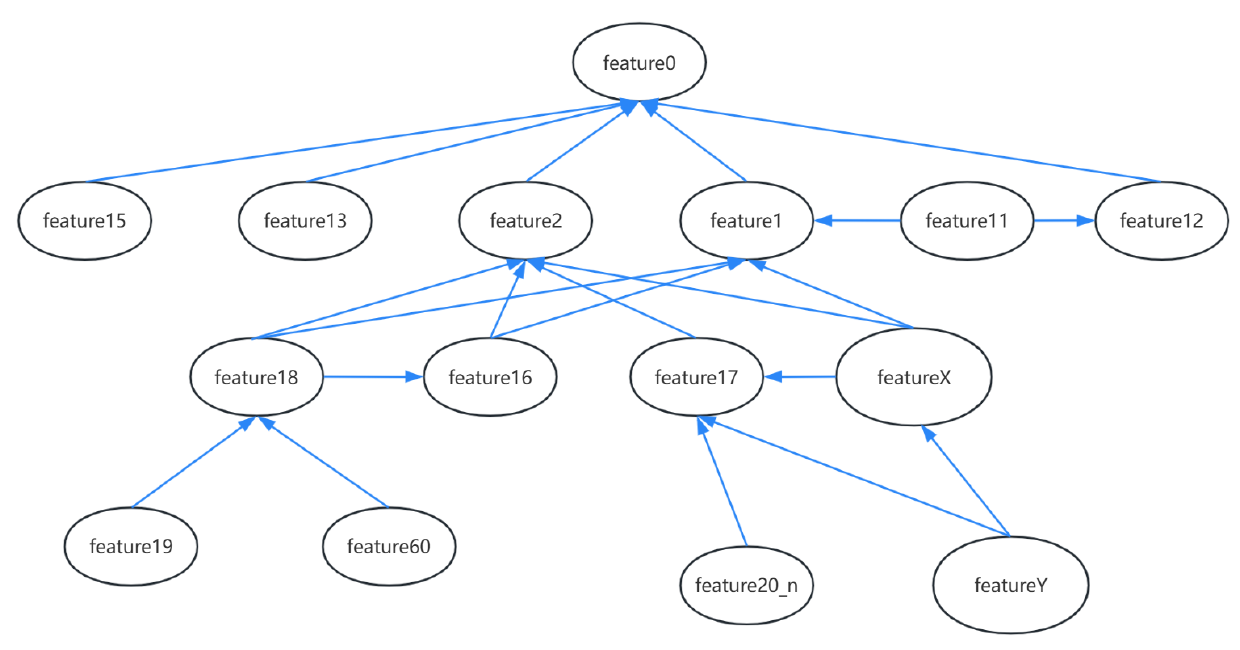}
    \caption{An incomplete causal relationship diagram after removing hidden variables summarized by some field experts.}
    \label{fig:subfig77}
\end{figure}
\begin{figure}[htbp]
    \centering
    \includegraphics[width=0.5\textwidth]{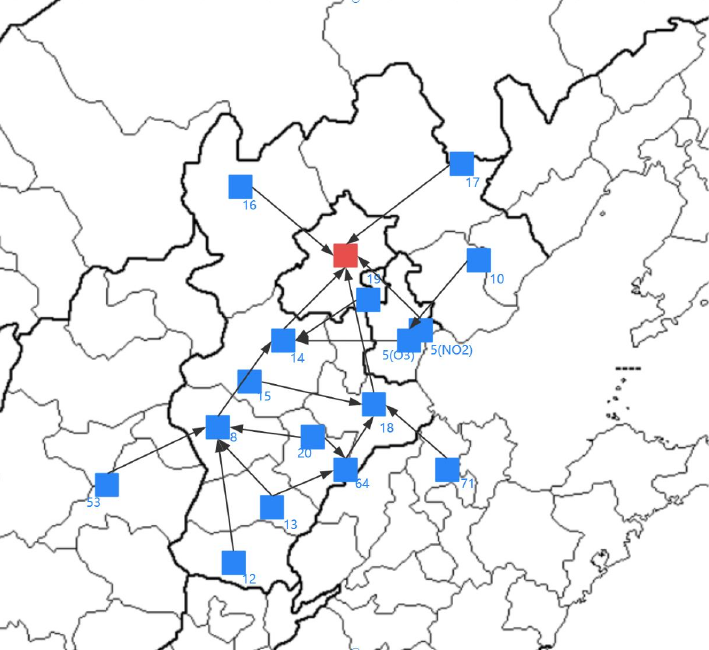}
    \caption{Causes of PM2.5 in Beijing before APEC}
    \label{fig:subfig8}
\end{figure}
\subsection{Experiment Setup}
\paragraph{Dataset} Experiments were conducted on both a simulated dataset based on random causal graphs and three real-world datasets. For the simulated dataset, SCMs were constructed using randomly generated causal graphs with \( n = 5, 10, 20 \) variables. The graph structures included both chain and random topologies, with sparsity levels set to \( 0.2 \) and \( 0.3 \). The weights of the causal edges were sampled to reflect varying degrees of correlation strength, including weak, medium, and strong associations. 
Training samples were generated according to Definition \ref{scm}, and anomalies were introduced by perturbing the noise terms of randomly selected variables. For each configuration, five datasets were generated using different random seeds. All reported results are averaged over these five datasets, and partial standard deviations are reported to reflect variability across runs. 

In addition to the synthetic data, three real-world datasets were used: 
\begin{itemize}
    \item \textbf{AIOPs datasets } \cite{zhang2022icassp} The dataset contains a real 5G dataset, which contains a causal graph and a related feature dataset. The feature dataset contains 23 observable variables and about 45\% of the samples are carefully labeled with relevant root causes. Since the MiCCD method involving hidden variables remains under development, only observed variables are considered in the relationship graph. Figure \ref{fig:subfig77} shows the adjusted causal graph used in our experiments.
    
    \item \textbf{Lemma-RCA Dataset} \cite{zheng2024lemma} This dataset comprises a range of real system failures, encompassing practical application scenarios such as microservices and water treatment or distribution systems. Pre-processed IT system data were selected for experimentation.
    
    \item \textbf{Air Pollutants with Urban Dataset} \cite{zhu2018p} The dataset comprises air quality monitoring data collected from the Beijing Olympic Sports Center station during 2013–2014, as released by the China National Environmental Monitoring Center. It nncludes hourly concentration levels of pollutants such as PM2.5, PM10, SO2, and NO2. Zhu et al. examined various air quality protection measures implemented during the APEC meeting, including odd-even vehicle restrictions, industrial shutdowns, and construction halts, and evaluated their direct impact on air quality improvement.
    
    Figure \ref{fig:subfig8} illustrates the pollution relationships during the same quarter, presenting the causal pathways of the pollutants both prior to and during the APEC meeting. 
\end{itemize}

\paragraph{Evaluation Metric}
For all experimental evaluations, the following criteria are adopted:
\begin{itemize}
    \item \textbf{F1-score} \cite{powers2020evaluation} defined as the harmonic mean of precision and recall, this metric evaluates the correctness of decision-making.
    \item \textbf{Normalized Cost (N-Cost)} a metric that normalizes the cost associated with each intervention decision, reflecting economic feasibility.
    \item \textbf{Normalized Discounted Cumulative Gain at K (nDCG@k)} \cite{yining2013theoretical} measures the quality of ranking by quantifying the alignment between predicted root causes and their ideal ranks within the top-K positions.
    \item \textbf{Relative Mean Squared Error (r-MSE)} \cite{legendre1806nouvelles} normalizes the standard Mean Squared Error to assess the reliability of counterfactual predictions.
\end{itemize}
These metrics jointly assess the proposed framework from multiple dimensions. The F1-score ensures accuracy, N-Cost emphasizes cost-efficiency, nDCG@k captures the practical relevance of root cause rankings, and r-MSE verifies the robustness of counterfactual reasoning. This comprehensive evaluation is essential for real-world deployment, where both technical performance and operational feasibility must be considered.

\paragraph{Baselines}
To evaluate the effectiveness of the proposed method, comparisons are conducted against several baseline approaches described in the Related Work section, including NaiveRCA \cite{casalicchio2019visualizing}, CausalRCA \cite{budhathoki2022causal}, BIGEN \cite{nguyen2024root}, LIME \cite{ribeiro2018anchors}, LC \cite{ide2021anomaly}, AUF-MICNS \cite{duavoiding}. For the sake of fairness, the same cost function is employed for both the proposed method and AUF-MICNS. For detailed information regarding the baseline methods, please refer to \ref{app:baseline}.

\subsection{Results and Discussion}
\begin{figure*}[ht]
    \centering
        \begin{minipage}[t]{\textwidth}
            \centering
            \includegraphics[height=1.1cm,width=1\textwidth]{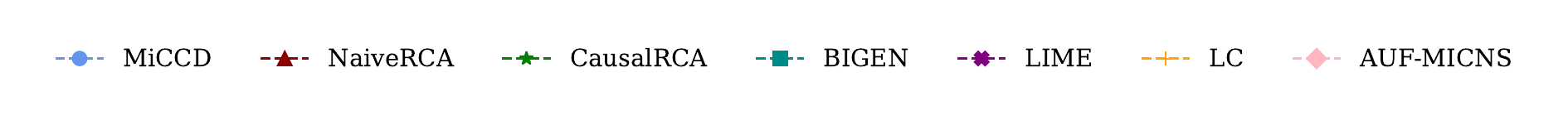}
        
        \end{minipage}
        
    \subfigure[Sensitivity of different number of nodes.]{
        \begin{minipage}[t]{0.3\textwidth}
        
            \centering
            \includegraphics[height=3cm,width=\textwidth]{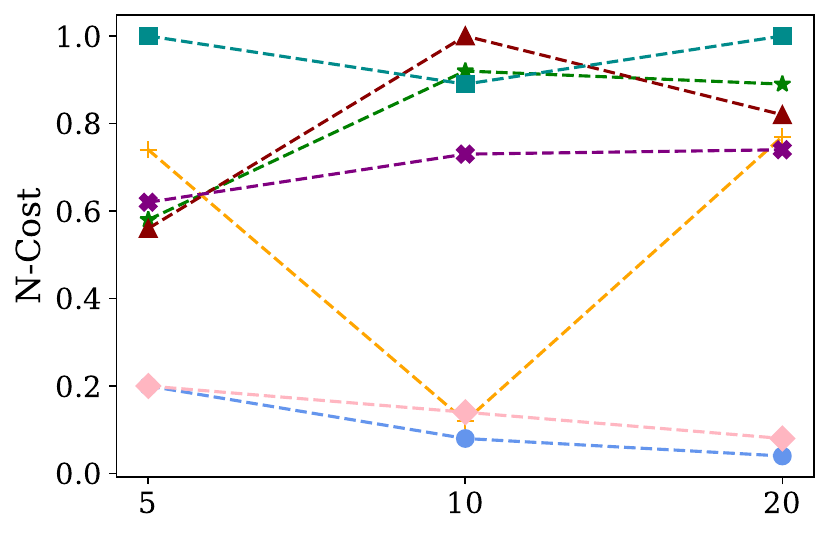}
            \includegraphics[height=3cm,width=\textwidth]{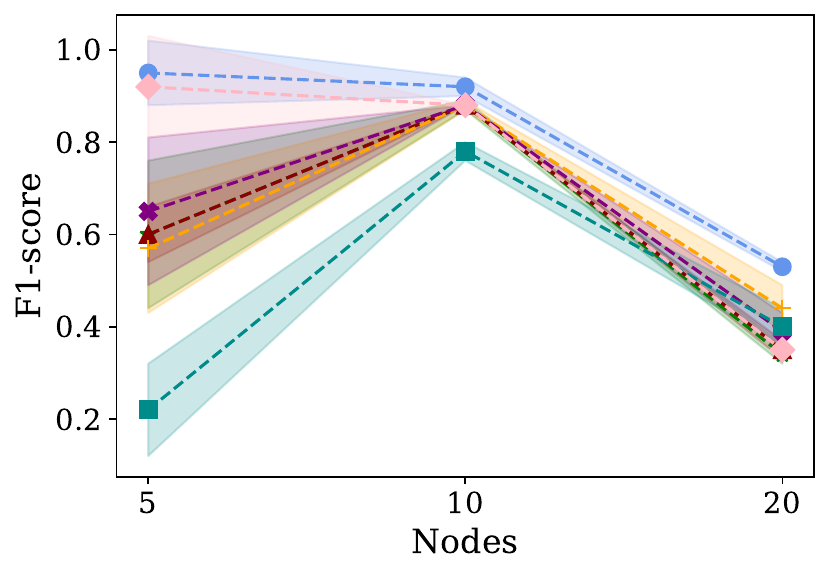}
        \end{minipage}
    }
    \subfigure[Sensitivity to different causal graph structures.]{
        \begin{minipage}[t]{0.3\textwidth}
            \centering
            \includegraphics[height=3cm,width=\textwidth]{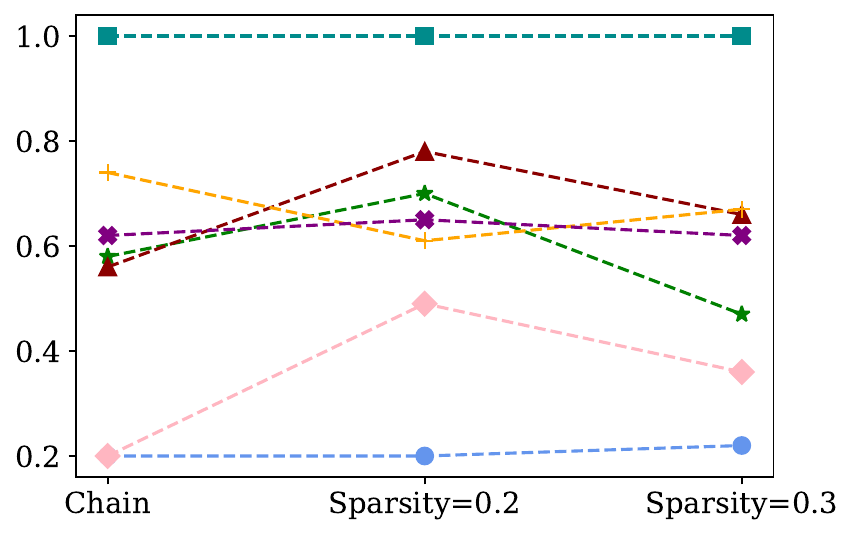}
            \includegraphics[height=3cm,width=\textwidth]{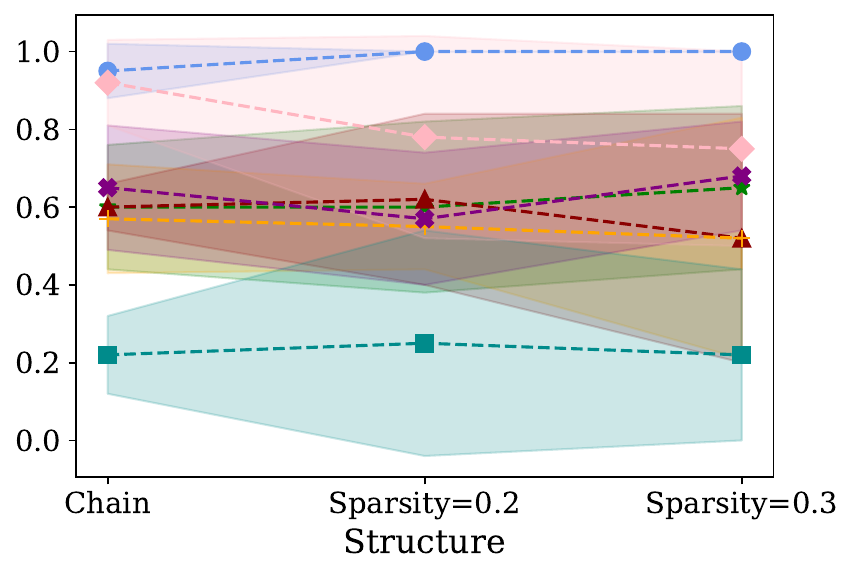}
        \end{minipage}
    }
    \subfigure[Sensitivity to different causal edge weight.]{
        \begin{minipage}[t]{0.3\textwidth}
            \centering
            \includegraphics[height=3cm,width=\textwidth]{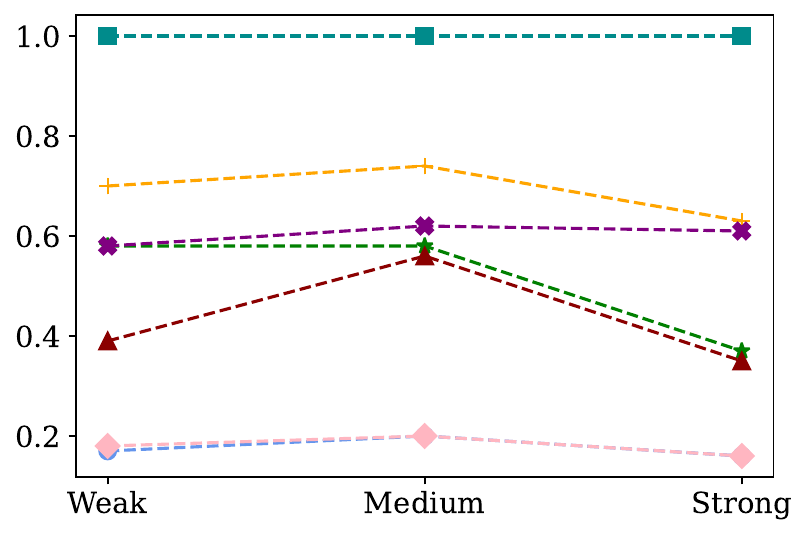}
            \includegraphics[height=3cm,width=\textwidth]{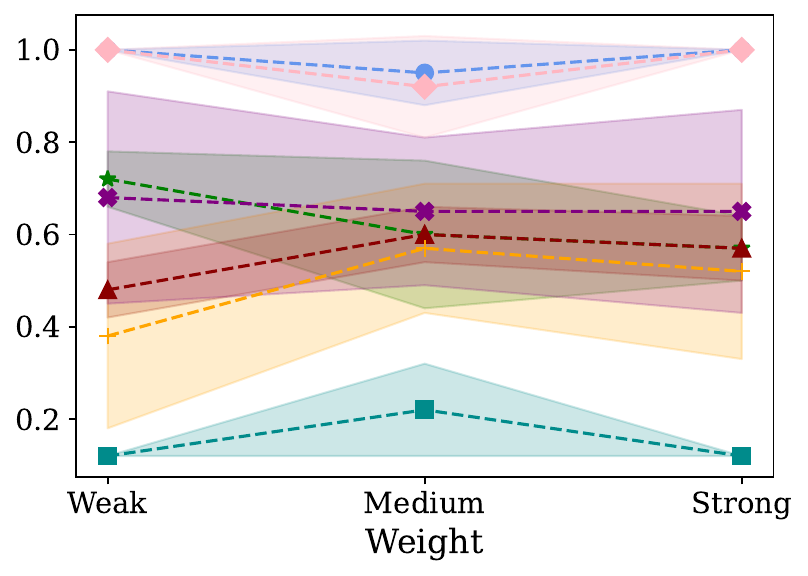}
        \end{minipage}
    }
    \vspace{-3mm}
    \caption{\textbf{Comparison of Results under Varying Experimental Settings.} The performance of different methods is evaluated under three varying settings:  the number of nodes (left), the sparsity level of the causal graph (middle), and the strength of causal edge weights (right). In each subfigure, the top row reports the normalized decision cost, while the bottom row presents the corresponding F1-score. It can be observed that the proposed method consistently achieves the lowest intervention cost and yields superior F1-scores across all settings, indicating both higher decision efficiency and greater diagnostic accuracy compared to baseline approaches.}
    \label{fig:combined}
\end{figure*}

\begin{figure*}[ht]
    \centering
        \begin{minipage}[t]{\textwidth}
            \centering
            \includegraphics[height=1.1cm,width=1\textwidth]{figures/legend.pdf}
        
        \end{minipage}
        
    \subfigure[Nodes = 5.]{
        \begin{minipage}[t]{0.3\textwidth}
        
            \centering
            \includegraphics[height=3cm,width=\textwidth]{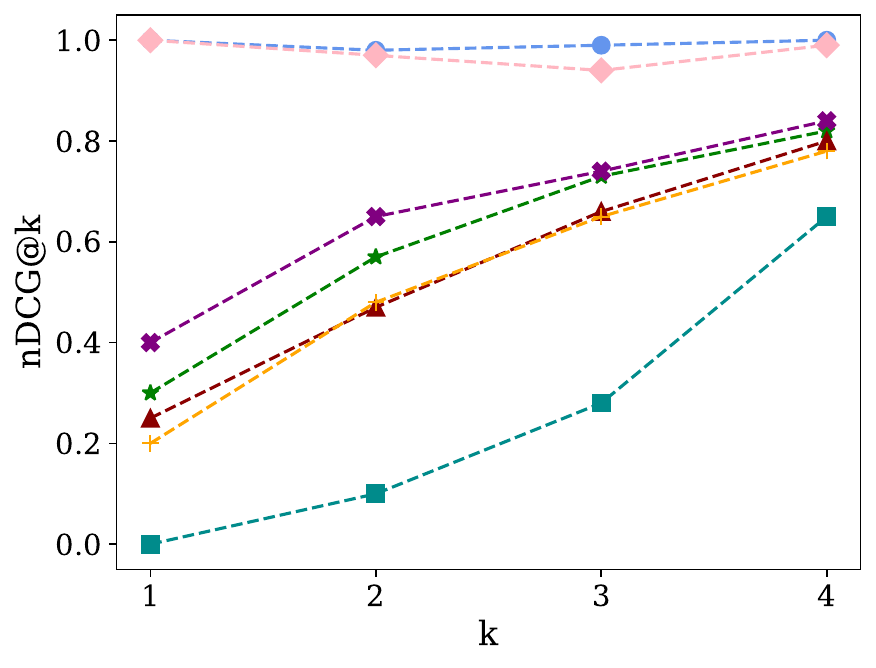}
        \end{minipage}
    }
    \subfigure[Nodes = 10.]{
        \begin{minipage}[t]{0.3\textwidth}
            \centering
            \includegraphics[height=3cm,width=\textwidth]{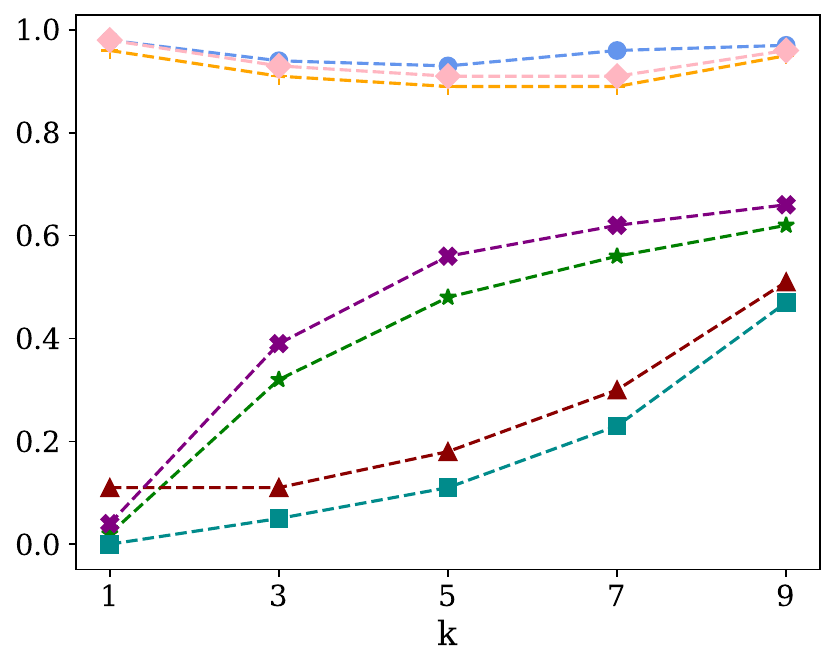}
        \end{minipage}
    }
    \subfigure[Nodes = 20.]{
        \begin{minipage}[t]{0.3\textwidth}
            \centering
            \includegraphics[height=3cm,width=\textwidth]{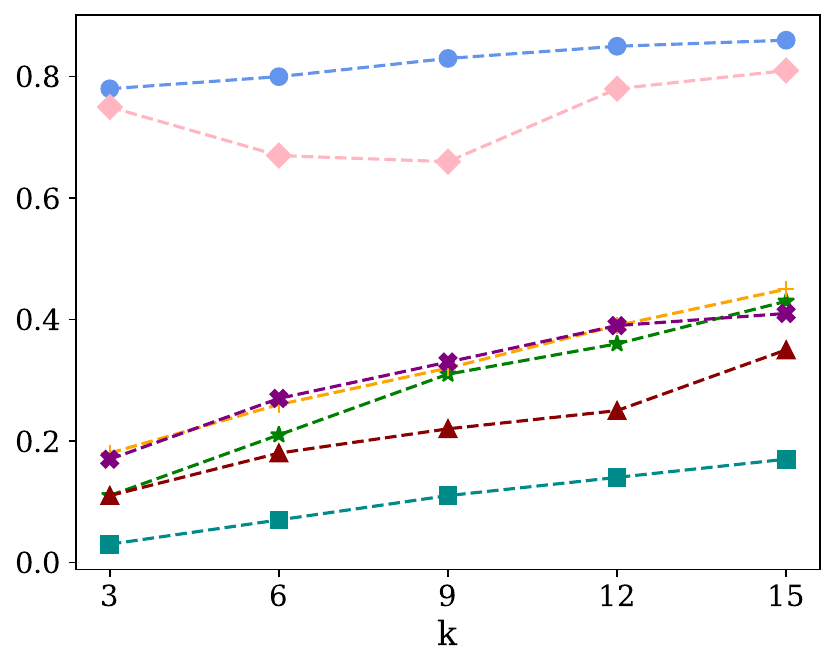}
        \end{minipage}
    }
    \vspace{-3mm}
    \caption{\textbf{Comparison of nDCG@k Values under Varying Numbers of Nodes.} The nDCG@k values are compared across different values of $k$ and varying numbers of nodes. Across all configurations, MiCCD consistently yields higher nDCG@k scores relative to baseline approaches, indicating improved ranking quality and more effective identification of root causes.}
    \label{fig:ndcg}
\end{figure*}

\paragraph{Overall Performance (Q1)} 
The N-Cost, F1-score, and nDCG@k values of all methods are compared across multiple simulated datasets. The results are summarized in Figure \ref{fig:combined} and Figure \ref{fig:ndcg}, highlighting the overall performance across varying experimental conditions. As illustrated in the figures, the proposed method consistently outperforms all baselines in terms of cost-efficiency, decision accuracy, and ranking quality.

In Figure \ref{fig:combined}, each column represents experiments conducted under different controlled settings. The first column reports results under varying numbers of nodes; the second column reflects the impact of different causal graph structures, characterized by varying levels of sparsity; and the third column presents performance under different strengths of causal edge weights. In each experiment, a single factor is varied while the other two are held constant, allowing for a clear assessment of its individual influence on model performance.

As illustrated in Figure \ref{fig:combined}, the first row of results indicates that MiCCD consistently achieves the lowest decision cost across all experimental settings. It is important to emphasize that the same cost evaluation metric is employed for both MiCCD and AUF-MICNS. While AUF-MICNS demonstrates lower decision costs than other baseline methods (excluding MiCCD), its cost remains consistently higher than that of MiCCD. 
The second row of results demonstrates that MiCCD not only achieves the highest F1-score under varying conditions but also exhibits significantly narrower error bars compared to competing approaches. These findings indicate that MiCCD is capable of reducing both false alarm rates and decision-making costs, while maintaining a balanced trade-off between precision and recall, thereby confirming its superior overall performance. 
Furthermore, as the number of nodes and structural complexity of the causal graph increase, MiCCD maintains strong stability and adaptability--particularly in random graphs with higher sparsity. These results underscore the applicability of MiCCD in real-world settings characterized by limited intervention resources and high costs associated with false alarms, highlighting its potential in supporting economically efficient and robust decision-making across a wide range of practical domains.

Figure \ref{fig:ndcg} presents the nDCG performance of all methods for different values of $k$ across datasets with 5, 10, and 20 nodes.

The performance of CausalRCA, NaiveRCA, LIME, and LC exhibits a strong dependence on the value of $k$. When $k$ is small, their nCDG values remain low, indicating limited ranking ability. In contrast, both MiCCD and AUF-MICNS demonstrate relative stability across varying conditions. However, AUF-MICNS exhibits more pronounced fluctuations, highlighting that MiCCD consistently performs well across generalization tasks. This stability is crucial for decision-making processes, ensuring the accurate and reliable prioritization of potential solutions.

\paragraph{Performance RCA based on Effective Interventions (Q2)}

As quantitatively shown in Table \ref{table:real}, MiCCD achieves state-of-the-art performance across all three real-world datasets. This sustained advantage offers two key insights into the MiCCD in root cause localization:

Advantages in Complex Scenarios: 
To demonstrate MiCCD's capability in guiding root cause localization, potential root causes are treated as options for resolving data anomalies, and the problem is formulated as a multi-class classification task. The F1-score is computed to assess whether the intervention variables identified by the proposed method correspond to the true root causes. In this tasks (AIOP and Lemma-RCA), MiCCD outperforms specialized baselines by up to 44 percentage points (pp). Notably, even specialized root cause localization methods, such as NaiveRCA, CausalRCA, and BIGEN, do not surpass MiCCD in this classic scenario. MiCCD outperforms causal Shapley-value-based methods(CausalRCA) by 23-46pp, highlighting its advantage in decision-making by minimizing cost. Compared to robust correlation-based methods(LIME), MiCCD still maintains a lead of 12-44pp, demonstrating its ability to avoid spurious dependencies.

Binary task efficiency: To evaluate performance-oriented root cause analysis based on effective interventions, the Air Pollutants with Urban Dataset was utilized under the assumption that certain pollutants were unaffected by the interventions implemented during the APEC period and were thus labeled as non-intervention variables, whereas those exhibiting significant reductions were classified as intervention variables. This setup enables the task to be framed as a binary classification problem, in which the objective is to distinguish between intervenable and non-intervenable variables using two real-world datasets. The F1-score is computed to evaluate the alignment of the identified variables with ground-truth intervention outcomes, thereby reflecting the effectiveness of the proposed method and baseline algorithms in actionable decision-making scenarios. Although both MiCCD and BIGEN achieved perfect scores on the air pollutant dataset, the simplicity of the task—focused solely on identifying intervenability—limits its capacity to reveal distinctions in intervention planning performance. Importantly, MiCCD further demonstrates its advantage in performance-oriented root cause analysis by incorporating minimum-cost optimization, thereby enabling the generation of equally accurate yet more efficient intervention strategies compared to BIGEN.

\begin{table}[ht]
\centering
\caption{F1-score of different methods on three real dataset tasks}
\begin{tabular}{cccc}
\toprule
Method & AIOPs & Air Pollutants & Lemma-RCA\\
\midrule
MiCCD & \textbf{0.95} & \textbf{1.00} &\textbf{0.94} \\
NaiveRca & 0.79 & 0.76 & 0.44\\
CausalRca & 0.72 & 0.80 &0.48\\
BIGEN & 0.88 & \textbf{1.00} &0.63\\
LIME & 0.83 & 0.74 & 0.50\\
LC & 0.87 & 0.82 & 0.50\\
AUF-MICNS & 0.82 & 0.60 & 0.50\\
\bottomrule
\end{tabular}
\label{table:real}
\end{table}

\begin{figure}[h] 
    \centering    
    \includegraphics[width=0.5\textwidth]{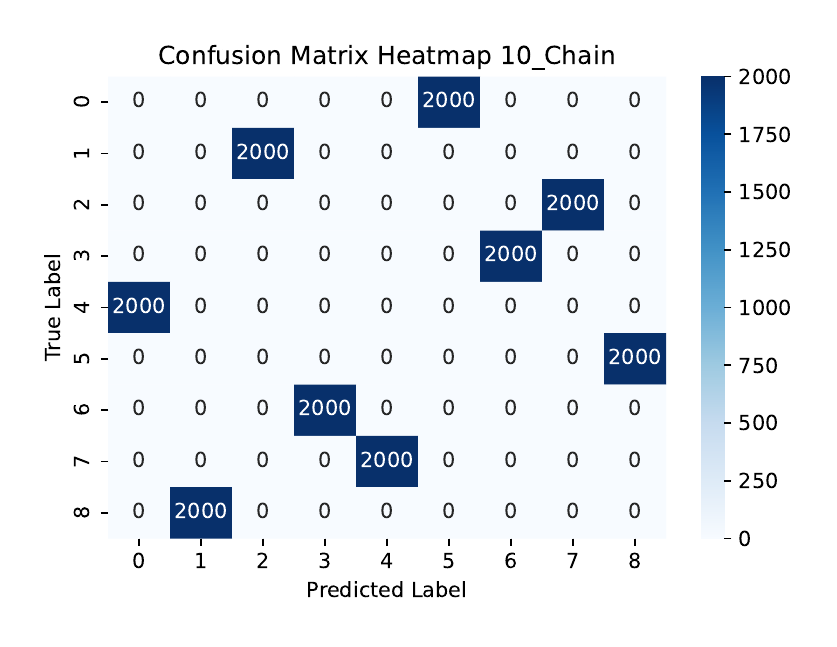} 
    \caption{Confusion matrix heat map: Clustering results when the dataset contains 10 nodes, a chain causal structure, and the causal edge weights are at a medium level.}
    \label{fig:cm}
\end{figure}

\paragraph{Component Validity and Counterfactual Capability (Q3)}

To evaluate the importance of each component in the surrogate model, two variants were devised as follows: a) \textbf{MiCCD-u}, which excludes abnormal pattern labels used as supervisory signals; and b) \textbf{MiCCD-g}, which omits the causal graph and bypasses variable-wise iterate. Table \ref{r-mse} presents the performance of MiCCD, its two variants, and the inferred model of AUF-MICNS \cite{duavoiding} in counterfactual estimation. exhibits superior performance with 10 nodes, attributed to the ease of learning causal mechanisms in chain structures, rendering the causal graph less critical. However, in more complex structures (e.g., sparsity = 0.3), MiCCD-g's r-MSE increases from 0.011 to 0.038, underscoring the value of incorporating the causal graph. The results further indicate the critical role of pattern-label supervision, particularly in low-dimensional settings. Notably, AUF-MICNS exhibits consistently poor performance due to its reliance on population-level optimization, which is unsuitable for instance-specific anomalies. While the proposed variants are limited in handling abnormal data, they demonstrate greater adaptability than AUF-MICNS, thereby supporting the robustness of MiCCD. Reconstruction performance is further validated in \ref{app:exp}.

Taking a dataset with 10 nodes, chain causal structure and medium causal weight as an example, Figure \ref{fig:cm} illustrates the effectiveness of using abnormal pattern clustering to distinguish abnormal patterns. This high separability provides a reliable supervisory signal; when removed (MiCCD-u), the counterfactual r-MSE increases across the entire structure, as shown in Table \ref{r-mse}.

\begin{table}[h]
\caption{Counterfactual Estimation Average r-MSE of the three methods under different settings}
\label{r-mse}

\centering
\setlength{\tabcolsep}{3mm}
\renewcommand{\arraystretch}{1.0}
\normalsize
\begin{tabular}{ccccc}
\toprule
Structure & Method & 5 & 10 & 20 \\
\midrule
\multirow{4}{*}{Chain} & MiCCD & \textbf{0.019} & 0.018 & \textbf{0.008} \\
& MiCCD-u & 0.040 & 0.037 & 0.009 \\
& MiCCD-g & 0.050 & \textbf{0.011} & 0.048 \\
& AUF-MICNS & 0.168 & 0.098 & 0.059 \\
\midrule
\multirow{4}{*}{Sparsity=0.2} &MiCCD & \textbf{0.063} & \textbf{0.043} & \textbf{0.018} \\
&MiCCD-u & 0.067 & 0.069 & 0.031 \\
& MiCCD-g & 0.117 & 0.063 & 0.084 \\
& AUF-MICNS & 0.251 & 0.196 & 0.173 \\
\midrule
\multirow{4}{*}{Sparsity=0.3} & MiCCD & \textbf{0.038} & \textbf{0.020} & \textbf{0.004} \\
&MiCCD-u & 0.100 & 0.034 & 0.005 \\
&MiCCD-g & 0.079 & 0.038 & 0.007 \\
&AUF-MICNS & 0.206 & 0.131 & 0.035 \\
\bottomrule
\end{tabular}
\end{table}

\section{Conclusion}
The Minimum Cost Causal Decision (MiCCD) framework is introduced as an innovative paradigm for decision-making in complex anomaly-emerging scenarios, including AI operations and maintenance. MiCCD integrates causal reasoning with continuous optimization to address the challenge of minimizing intervention costs while achieving desired outcomes. The framework comprises two principal components: (a) a surrogate model that approximates real-world causal relationships through cluster labels of abnormal patterns; and (b) an optimization strategy that employs counterfactual estimation to identify the most cost-effective interventions. Empirical results demonstrate that MiCCD exhibits adaptability across diverse decision-making contexts by assigning variable-specific costs, thereby enabling both root cause identification and targeted interventions. The framework supports personalized decision-making by tailoring strategies to specific systems or scenarios, (e.g., personalized medicine or smart manufacturing), thereby highlighting its broad applicability and effectiveness.

\appendix
\section{Evidence Lower Bound}\label{app:elbo}
In this subsection, we show the evidence lower bound.
\begin{equation}
\begin{aligned}
    \log p(\mathbf{x} |\mathbf{u})
=&\log p(x_1,x_2,\dots ,x_d|\mathbf{u} )
\\=&\sum_{j=1}^{d} \log_p(x_j|\mathbf{x}_{PA_{j}^{\mathcal{G} }},\mathbf{u}  )
\\=&\sum_{j=1}^{d} \log\int p(x_j,z_j|\mathbf{x}_{PA_{j}^{\mathcal{G} }},\mathbf{u} )dz_j
\\=&\sum_{j=1}^{d} \log\int p(x_j|z_j,\mathbf{x}_{PA_{j}^{\mathcal{G} }},\mathbf{u} )p(z_j|\mathbf{x}_{PA_{j}^{\mathcal{G} }},\mathbf{u} )dz_j
\\=&\sum_{j=1}^{d} \log\int p(x_j|z_j,\mathbf{x}_{PA_{j}^{\mathcal{G} }},\mathbf{u} )p(z_j|\mathbf{x}_{PA_{j}^{\mathcal{G} }},\mathbf{u} )\frac{q(z_j|\mathbf{x}_{PA_{j}^{\mathcal{G} }},x_j,\mathbf{u})}{q(z_j|\mathbf{x}_{PA_{j}^{\mathcal{G} }},x_j,\mathbf{u})}dz_j
\\=&\sum_{j=1}^{d} \log  \mathbb{E} _{q(z_j|\mathbf{x}_{PA_{j}^{\mathcal{G} }},x_j,\mathbf{u} )}\left [ \frac{p(x_j|z_j,\mathbf{x}_{PA_{j}^{\mathcal{G} }},\mathbf{u} )p(z_j|\mathbf{x}_{PA_{j}^{\mathcal{G} }},\mathbf{u} )}{q(z_j|\mathbf{x}_{PA_{j}^{\mathcal{G} }},x_j,\mathbf{u} )}  \right ] 
\\\ge &\sum_{j=1}^{d}\mathbb{E} _{q(z_j|\mathbf{x}_{PA_{j}^{\mathcal{G} }},x_j,\mathbf{u} )}\left [ \log \frac{p(x_j|z_j,\mathbf{x}_{PA_{j}^{\mathcal{G} }},\mathbf{u} )p(z_j|\mathbf{x}_{PA_{j}^{\mathcal{G} }},\mathbf{u} )}{q(z_j|\mathbf{x}_{PA_{j}^{\mathcal{G} }},x_j,\mathbf{u} )}  \right ] 
\\=&\sum_{j=1}^{d}\mathbb{E} _{q(z_j|\mathbf{x}_{PA_{j}^{\mathcal{G} }},x_j,\mathbf{u} )}\left [\log p(x_j|z_j,\mathbf{x}_{PA_{j}^{\mathcal{G} }},\mathbf{u} ) +\log p(z_j|\mathbf{x}_{PA_{j}^{\mathcal{G} }},\mathbf{u} ) -\log q(z_j|\mathbf{x}_{PA_{j}^{\mathcal{G} }},x_j,\mathbf{u} )   \right ]
\\=& \sum_{j=1}^{d}\mathbb{E} _{q(z_j|\mathbf{x}_{PA_{j}^{\mathcal{G} }},x_j,\mathbf{u} )}\left [ \log p(x_j|z_j,\mathbf{x}_{PA_{j}^{\mathcal{G} }},\mathbf{u} ) \right ] \\
&-\sum_{j=1}^{d} D_{KL}( q(z_j|\mathbf{x}_{PA_{j}^{\mathcal{G} }},x_j,\mathbf{u} )\parallel p(z_j|\mathbf{x}_{PA_{j}^{\mathcal{G} }},\mathbf{u}))
\end{aligned}
\end{equation}

\begin{table*}[]
\caption{MLP Training Configuration(We use AP for Air Pollutants with Urban)}
\label{config}
\centering
\setlength{\tabcolsep}{3mm}{
\resizebox{\textwidth}{!}{
\renewcommand{\arraystretch}{0.7}
\begin{tabular}{c|cccccccccccc}
\toprule
                    & \multicolumn{3}{c|}{\textbf{Chain}}                       & \multicolumn{3}{c|}{\textbf{Sparsity=0.2}}                      & \multicolumn{3}{c|}{\textbf{Sparsity=0.3}}             \\ 
                    \midrule
\textbf{Dataset}      & 5          &10              & \multicolumn{1}{c|}{20}      & 5          &10              & \multicolumn{1}{c|}{20}& 5          &10              & \multicolumn{1}{c|}{20}  &\multicolumn{1}{c|}{AIOPs} &\multicolumn{1}{c|}{AP}&Lemma-RCA\\ \midrule
\textbf{Batch Size}     & 64   &64        & \multicolumn{1}{c|}{64}  & 64    &64      & \multicolumn{1}{c|}{64} & 64     &64     & \multicolumn{1}{c|}{64} &\multicolumn{1}{c|}{64} &\multicolumn{1}{c|}{64}&64   \\
\textbf{Epoch}      & 20   & 20       & \multicolumn{1}{c|}{20}  & 20     &20     & \multicolumn{1}{c|}{20} & 20     & 20    & \multicolumn{1}{c|}{20} &\multicolumn{1}{c|}{20} &\multicolumn{1}{c|}{20}&20  \\
\textbf{Hidden Dim}    & 50     &30     & \multicolumn{1}{c|}{30}  & 50     &30     & \multicolumn{1}{c|}{30} & 50     &30     & \multicolumn{1}{c|}{30} & \multicolumn{1}{c|}{50}& \multicolumn{1}{c|}{50}&50 \\
\textbf{Depth}      & 3   & 3       & \multicolumn{1}{c|}{3}  & 3     &3     & \multicolumn{1}{c|}{3} & 3     & 3    & \multicolumn{1}{c|}{3} &\multicolumn{1}{c|}{3} &\multicolumn{1}{c|}{3} &3  \\
\textbf{Learning Rate}      & 1e-3   & 1e-3       & \multicolumn{1}{c|}{1e-3}  & 1e-3     &1e-3     & \multicolumn{1}{c|}{1e-3} & 1e-3     & 1e-3    & \multicolumn{1}{c|}{1e-3} &\multicolumn{1}{c|}{1e-3} &\multicolumn{1}{c|}{1e-3} &1e-3  \\
\textbf{Activation}      & lrelu   & lrelu       & \multicolumn{1}{c|}{lrelu}  & lrelu     &lrelu     & \multicolumn{1}{c|}{lrelu} & lrelu     & lrelu    & \multicolumn{1}{c|}{lrelu} &\multicolumn{1}{c|}{lrelu} &\multicolumn{1}{c|}{lrelu} &lrelu 
\\  \bottomrule
\end{tabular}}
}
\end{table*}

\section{Baseline Details}
\label{app:baseline}
\begin{itemize}
    \item \textbf{NaiveRCA} \cite{casalicchio2019visualizing} ranks variables by directly applying existing outlier scores, such as z-scores, to estimate their likelihood of being the root cause of anomalies.
    \item \textbf{CausalRCA} \cite{budhathoki2022causal} employs a functional causal model (FCM) to quantify the contribution of each variable observed anomalies. Shapley-value are utilized to fairly evaluate these contributions, thereby facilitating the formal identification of root causes.
    \item \textbf{BIGEN} \cite{nguyen2024root} prosents a framework that integrates Bayesian inference with gradient-based attribution methods to effectively identify and attribute the root causes of anomalies in complex systems, while accounting for both node and edge noise.
    \item \textbf{LIME} \cite{ribeiro2018anchors} constructs a local neighborhood around a test sample by sampling nearby data points. A linear model, such as linear regression, is then trained to fit this neighborhood, and the prediction is interpreted through the model’s coefficients, which reflect the contribution of each feature to the outcome.
    \item \textbf{LC} \cite{ide2021anomaly} assigns a responsibility score for each input variable by assessing the deviation between the model prediction and the actual observed value. A correction term  is derived based on the likelihood to explain anomalous predictions.
    \item \textbf{AUF-MICNS} \cite{duavoiding} update the influence relationship between variables dynamically via online gradient descent and integration techniques. It subsequently generates minimum-cost change recommendations through probability region construction and convex quadratic constrained quadratic programming (QCQP). 
\end{itemize}

\section{Implementation Details}
\label{app:a5}
The training configuration of the multilayer perceptron (MLP) models used in both the encoder and decoder is provided in Table \ref{config}.

\section{More Experiment Results}
\label{app:exp}
Thr reconstruction results of the model on the test set are presented. Table \ref{re-mse} summarizes the r-MSE of MiCCD, MiCCD-u, and AUF-MICNS on the simulated dataset. Since MiCCD-g provides the complete input vector $\mathbf{x}$ to the decoder during training, it is excluded from the evaluation. The results indicate that the proposed method significantly outperforms both its variant and the baseline.
\begin{table*}[]
\caption{Reconstruction Average relative MSE of the three methods under different settings}
\label{re-mse}
\centering
\setlength{\tabcolsep}{3mm}{
\resizebox{\textwidth}{!}{
\renewcommand{\arraystretch}{0.7}
\begin{tabular}{c|ccccccccc}
\toprule
                    & \multicolumn{3}{c|}{\textbf{Chain}}                       & \multicolumn{3}{c|}{\textbf{Sparsity=0.2}}                      & \multicolumn{3}{c}{\textbf{Sparsity=0.3}}             \\ 
                    \midrule
\textbf{Methods}      & 5          &10              & \multicolumn{1}{c|}{20}      & 5          &10              & \multicolumn{1}{c|}{20}& 5          &10              & 20\\ \midrule
\textbf{MiCCD}     & \textbf{0.025}   &\textbf{0.017}        & \multicolumn{1}{c|}{\textbf{0.008}}  & \textbf{0.036}    &\textbf{0.023}      & \multicolumn{1}{c|}{\textbf{0.013}} & \textbf{0.018}     &\textbf{0.011}     & \multicolumn{1}{c}{\textbf{0.002}}    \\
\textbf{MiCCD-u}      & 0.034   & 0.034       & \multicolumn{1}{c|}{0.009}  & 0.039     &0.042     & \multicolumn{1}{c|}{0.016} & 0.070     & 0.027    & \multicolumn{1}{c}{0.003}   \\
\textbf{AUF-MICNS}    & 0.135     &0.088     & \multicolumn{1}{c|}{0.057}  & 0.211      & 0.156    & \multicolumn{1}{c|}{0.134} & 0.151     & 0.109     & \multicolumn{1}{c}{0.027} 
\\  \bottomrule
\end{tabular}}
}
\end{table*}

\bibliographystyle{elsarticle-harv}
\bibliography{reference}

\end{document}